\documentclass{article}

\PassOptionsToPackage{numbers, compress}{natbib}

\usepackage[preprint]{neurips_2026}

\usepackage[utf8]{inputenc} 
\usepackage[T1]{fontenc}    
\usepackage{graphicx}       
\usepackage{hyperref}       
\usepackage{url}            
\usepackage{array}          
\usepackage{booktabs}       
\usepackage{multirow}       
\usepackage{algorithm}      
\usepackage{algpseudocode}  
\usepackage{amsmath}        
\usepackage{amsfonts}       
\usepackage{amssymb}        
\usepackage{nicefrac}       
\usepackage{microtype}      
\usepackage{xcolor}         

\newcommand{\xmark}{\ensuremath{\times}}

\title{Recurrent Residual Quantization: A Progressive Multi-Precision Representation for LLMs}

\author{%
  Yu Luo \\
  Intel \\
  \And
  Bo Dong \\
  Intel  \\
  \And
  Wenhua Cheng \\
  Intel  \\
  \And
  Haihao Shen \\
  Intel  \\
}

\begin{document}

\maketitle

\begin{abstract}
Serving large language models (LLMs) under diverse deployment constraints requires flexible trade-offs between accuracy, memory footprint, and throughput. However, conventional quantization methods typically require a separate checkpoint for each target bit-width. We introduce Recurrent Residual Quantization (RRQ), a post-training quantization (PTQ) framework that represents weights as a low-bit quantized base together with a sequence of quantized residual corrections, enabling multiple effective precisions from a single checkpoint. Starting from a 2-bit model obtained via post-training quantization (PTQ) or round-to-nearest (RTN), RRQ progressively adds lightweight 2-bit residuals generated via RTN to construct 4-, 6-, and 8-bit representations. The method is calibration-free and avoids joint multi-bit optimization. In our Qwen3-8B setup, the full all-RTN 2-/4-/6-/8-bit package is constructed in 1,293 seconds, $3.3\times$ faster than the measured MatGPTQ construction. Experiments on six recent LLMs show competitive accuracy at 6 and 8 bits, with model-dependent behavior at 4 bits. The code will be made publicly available upon publication.
\end{abstract}

\section{Introduction}

Large language models (LLMs)~\citep{achiam2023gpt,liu2024deepseek,qwen3,llama3} have achieved strong performance across a wide range of tasks, but their memory footprint and bandwidth demand remain major barriers to efficient serving. Weight-only post-training quantization (PTQ) is a practical way to reduce these costs, and recent methods can preserve strong accuracy at low precision~\citep{gptq,awq,quip,omniquant,autoround}. However, most PTQ pipelines are built for a single target precision: supporting 4-bit, 6-bit, and 8-bit deployment typically requires constructing and storing separate checkpoints.

This fixed-precision workflow is poorly matched to flexible serving. In practice, the preferred precision may depend on available memory, latency targets, workload size, and accuracy requirements~\citep{distserve,kwon2023efficient}. A single representation that exposes multiple accuracy--efficiency trade-offs would reduce checkpoint management overhead and avoid repeatedly quantizing the same model for different bit-widths. Recent single-checkpoint multi-precision methods~\citep{matquant,matgptq} address this goal by using nested integer bit layouts, where lower-precision models are derived from a shared higher-bit representation. While effective, this design couples all supported precisions within one bit hierarchy and makes it difficult to reuse an already optimized low-bit checkpoint as the starting point for higher-precision variants.

We introduce \emph{Recurrent Residual Quantization} (RRQ), a post-training framework for single-checkpoint multi-precision LLM representation. RRQ replaces nested bit slicing with additive residual refinement: weights are represented as a low-bit quantized base plus a sequence of quantized residual corrections. The base stage gives the lowest-precision model, and adding residual stages progressively improves the reconstruction. This formulation separates the base quantizer from the precision-expansion mechanism, allowing the low-bit foundation and the residual stages to be constructed with different quantizers.

Our main implementation uses RTN for the 2-bit base and for three lightweight 2-bit residual stages, yielding 2-, 4-, 6-, and 8-bit representations. The entire construction is calibration-free and does not require Hessian estimation or joint multi-bit optimization. As a result, RRQ can efficiently construct a multi-precision package without depending on a learned low-bit base quantizer. We also report a stronger SignRoundV2-base variant as an ablation to separate the effect of first-stage quality from the residual representation itself. In a representative Qwen3-8B case study, the all-RTN RRQ package completes in 1,293 seconds, about $3.3\times$ faster than a prior multi-precision PTQ baseline.

We further analyze when residual refinement is expected to help. The analysis shows that RRQ is most favorable when localized outliers dominate the dynamic range of quantization groups. In this regime, the base stage captures large-magnitude components, leaving later stages to quantize a narrower residual signal. This view also explains why RRQ's low-bit behavior is model-dependent: residual refinement is more effective for outlier-heavy weight distributions, while direct or specialized fixed-bit quantizers may remain preferable for flatter distributions.

We evaluate RRQ on six recent LLMs covering both base and instruction-tuned checkpoints. RRQ achieves near-BF16 accuracy at higher effective precisions and remains competitive with existing single-checkpoint multi-precision methods at 8 and 6 bits. At 4 bits, performance varies more across models, consistent with the outlier-based analysis.

Our contributions are summarized as follows:
\begin{itemize}
  \item We introduce RRQ, a post-training framework that represents LLM weights as a low-bit quantized base plus quantized residual stages, enabling single-checkpoint multi-precision reconstruction and reuse of existing low-bit checkpoints.

  \item We analyze RRQ through an outlier-based lens, showing when progressive residual refinement can be preferable to direct fixed-bit quantization and explaining its model-dependent behavior at low bit-widths.

  \item We validate RRQ on recent LLMs, showing competitive accuracy at 8 and 6 bits, model-dependent 4-bit behavior, and a measured $3.3\times$ construction-time reduction in a Qwen3-8B case study.
\end{itemize}

\section{Related Work}

\subsection{Fixed-Precision Weight Quantization for LLMs}

Weight quantization is a standard way to reduce the memory footprint and inference cost of LLMs. Post-training quantization (PTQ) compresses pretrained weights without retraining. Round-to-nearest (RTN) is a simple baseline, while GPTQ~\citep{gptq} and AWQ~\citep{awq} use second-order information or activation-aware scaling to reduce quantization loss. SqueezeLLM~\citep{squeezellm} uses non-uniform quantization for outlier-heavy weight distributions. Outliers were systematically highlighted by LLM.int8()~\citep{llmint8}, which showed that a small fraction of activation channels can dominate quantization error; SmoothQuant~\citep{smoothquant} mitigates this by shifting difficulty from activations to weights through equivalent per-channel scaling. Training-aware methods such as LLM-QAT~\citep{llmqat} incorporate quantization into training, while PTQ methods such as OmniQuant~\citep{omniquant} learn auxiliary scaling and shifting parameters by block-wise reconstruction. These methods are effective, but they produce \emph{one checkpoint per target precision}; serving multiple precisions therefore requires multiple independently quantized models.

\subsection{Multi-Precision Quantization}

A growing body of work aims to obtain a \emph{single} quantized model that can run at multiple precisions. MatQuant~\citep{matquant} introduces the Matryoshka idea for integer quantization: a high-bit parent model can be sliced by most-significant-bit extraction to produce lower-bit sub-models at inference time. However, MatQuant is tied to learning-based quantization (QAT or OmniQuant) and does not support one-shot PTQ or reuse of existing checkpoints.

MatGPTQ~\citep{matgptq} extends this line to PTQ by adapting GPTQ to a joint multi-bit objective, producing a sliceable parent checkpoint in one pass with heterogeneous per-layer bit allocation. Like MatQuant, however, MatGPTQ remains restricted to integer Matryoshka slicing and nested integer bit layouts.

Mixed-precision methods such as HAWQ~\citep{hawq}, HAWQ-V2~\citep{hawqv2}, and OWQ~\citep{owq} assign different bit-widths across layers, but they target one static precision profile rather than switching among multiple usable precisions.

RRQ is complementary to MatQuant and MatGPTQ. Instead of MSB slicing within one integer code, it decomposes weights into a base quantizer and recurrent residual stages. This preserves PTQ flexibility while removing the requirement that all precisions arise from nested integer bit fields. RRQ can therefore build on existing quantized checkpoints and reuse existing quantizers or low-bit kernels as stage-wise building blocks. It is also representationally compatible with heterogeneous stage formats, although this paper empirically evaluates only integer low-bit stages. Table~\ref{tab:method_comparison} summarizes the distinctions.

\begin{table}[t]
  \caption{Feature comparison with representative quantization approaches. \checkmark: supported and evaluated or standard for the method; repr.: representationally supported but not empirically evaluated here; \xmark: not supported; partial: restricted support; ---: outside the method's scope.}
  \label{tab:method_comparison}
  \centering
  \small
  \setlength{\tabcolsep}{3.5pt}
  \begin{tabular}{p{0.36\linewidth}cccc}
    \toprule
    & RRQ & GPTQ~\citep{gptq} & \shortstack[c]{MatQuant\\\citep{matquant}} & \shortstack[c]{MatGPTQ\\\citep{matgptq}} \\
    \midrule
    Single-checkpoint multi-precision & \checkmark & \xmark & \checkmark & \checkmark \\
    Post-training applicable & \checkmark & \checkmark & partial & \checkmark \\
    Non-integer / FP stages & repr. & \xmark & \xmark & \xmark \\
    Builds on quantized checkpoints & \checkmark & \xmark & \xmark & \xmark \\
    Reuses quantizers / kernels & \checkmark & --- & \xmark & \xmark \\
    \bottomrule
  \end{tabular}
\end{table}

\subsection{Residual and Multi-Stage Quantization}

Residual quantization has a long history in signal processing and vector quantization, where a signal is approximated by successively quantizing and subtracting reconstruction errors~\citep{juang1982}. In neural network compression, residual vector quantization (RVQ) has been used for codebook-based weight compression~\citep{martinez2021} and learned image compression~\citep{lee2022}.

Recent activation-compression work also uses residual refinement. Quant VideoGen (QVG)~\citep{qvg} applies progressive residual quantization to KV-cache tensors in auto-regressive video diffusion models, showing that iterative residual coding can reduce activation storage. RRQ differs in target and objective: it quantizes \emph{static model weights}, uses fixed per-group scalar quantization rather than input-dependent clustering, and makes every prefix of residual stages a usable model at a distinct effective bit-width.

\section{Recurrent Residual Quantization}
\label{sec:rrq-outlier-prevalence}

RRQ is motivated by the heavy-tailed weight distributions observed in modern LLMs. Prior studies report localized outliers in both activations~\citep{llmint8,massive_activations} and weights~\citep{squeezellm,owq,spqr} of transformer models. To quantify local outlier severity, we define the \textbf{Peak-to-Mean Ratio} (PMR) as the maximum absolute weight divided by the mean absolute value (MAE) within the same quantization group. Taking the maximum PMR across groups in a tensor highlights the most challenging groups for uniform quantization. For example, Qwen3-14B has a mean tensor-wise maximum PMR of 27.826 under group size 128. Such localized dynamic ranges can increase uniform quantization error, suggesting a setting in which residual correction may be useful.

\subsection{Problem Setup}

Let $x_i^j$ denote the $j$-th floating-point weight element within quantization group $i$, and let $s^i$ and $z^i$ be the shared scale and zero-point for that group. A conventional quantizer first applies an affine transformation to scale and shift each weight:
\begin{equation}
q^j = \frac{x_i^j}{s^i} + z^i,
\end{equation}
followed by a rounding operator $\mathcal{R}(\cdot)$ to obtain the discrete integer code:
\begin{equation}
Q^j = \mathcal{R}(q^j).
\end{equation}
The quantizer stores the integer code $Q^j$ alongside the metadata $s^i$ and $z^i$. Consequently, the quantization error, or residual, is defined as:
\begin{equation}
r^j = x_i^j - (Q^j - z^i)s^i.
\end{equation}

Standard fixed-bit quantization discards this residual, so each target precision is typically generated as an independent checkpoint. RRQ instead quantizes the residual $r^j$ in subsequent stages, producing a sequence of additive corrections.

\subsection{Methodological Framework}

Assume the base stage quantization format is denoted by $b_0$, which may correspond to an integer bit-width or a specialized low-bit floating-point format. We define $S$ as the total number of residual quantization stages, where $b_k$ represents the format allocated to the $k$-th residual stage. If all stages employ integer formats, the accumulated nominal bit budget after $t$ residual stages is given by $B_t=b_0+\sum_{k=1}^{t} b_k$. 

RRQ initiates from a base quantized model and recursively quantizes the residual errors:
\begin{align}
Q_0^j &= \mathcal{Q}_{b_0}(x^j, z_0^j), \\
r_0^j &= x^j - \hat{x}_0^j, \\
Q_k^j &= \mathcal{Q}_{b_k}(r_{k-1}^j, z_k^j), \quad \text{for } k = 1, \ldots, S, \\
r_k^j &= r_{k-1}^j - \hat{r}_k^j,
\end{align}
where $\hat{x}_0^j$ is the dequantized approximation from the base stage, and $\hat{r}_k^j$ denotes the dequantized correction from the $k$-th residual stage. The effective reconstructed weight after accumulating $t$ stages is simply the additive sum:
\begin{equation}
\tilde{x}^{j}_{(t)} = \hat{x}_0^j + \sum_{k=1}^{t} \hat{r}_k^j.
\end{equation}

Algorithm~\ref{alg:rrq} formalizes this construction. The protocol requires full-precision weights to compute residual targets. It can also start from an existing low-bit checkpoint by skipping the initial base quantization step.

\begin{algorithm}[t]
  \caption{Recurrent Residual Quantization (RRQ)}
  \label{alg:rrq}
  \begin{algorithmic}[1]
    \Require Full-precision weights $W$, and optionally an existing base checkpoint; base quantizer $\mathcal{Q}_0$; residual quantizers $\{\mathcal{Q}_k\}_{k=1}^{S}$.
    \If{a base checkpoint is provided}
      \State Load its codes, scales, and zero-points; dequantize to $\hat{W}_0$.
    \Else
      \State Quantize $W$ with $\mathcal{Q}_0$; store base codes, scales, and zero-points; dequantize to $\hat{W}_0$.
    \EndIf
    \State $R_0 \gets W - \hat{W}_0$
    \For{$k=1,\ldots,S$}
      \State Quantize $R_{k-1}$ with $\mathcal{Q}_k$; store residual codes, scales, and zero-points.
      \State Dequantize the residual stage to $\hat{R}_k$.
      \State $R_k \gets R_{k-1} - \hat{R}_k$
    \EndFor
    \State \Return stored stage codes, scales, and zero-points, with prefix-$t$ reconstruction $\tilde{W}_{(t)}=\hat{W}_0+\sum_{k=1}^{t}\hat{R}_k$.
  \end{algorithmic}
\end{algorithm}

When the base and all residual stages use a 2-bit format (e.g., a 2-bit base tensor paired with three 2-bit residual tensors), the representation supports 2-, 4-, 6-, and 8-bit operating points. The standalone 2-bit operating point uses only the base stage, while the 4-, 6-, and 8-bit operating points add one, two, and three residual stages, respectively. Although our experiments focus on low-bit integer stages, the RRQ formulation permits heterogeneous stage formats. The base stage can also be obtained from an independently quantized checkpoint.

The corresponding prefill and decoding computations follow the same stage-wise decomposition. Higher effective precisions can be evaluated by summing the outputs of the base and residual-stage GEMMs. Appendix~\ref{app:prefill-decode-computation} provides the arithmetic details.

\subsection{Numerical Example and Error Analysis}

To build intuition for when recurrent residual layering can reduce outlier-induced error relative to direct uniform quantization, we analyze a simple zero-point quantization example with an injected outlier. The inlier region is clamped to $[-0.5, 0.5]$ (so $r=0.5$), and a single outlier has magnitude $K$. We consider two settings: $K=5$ ($10r$) and $K=3$ ($6r$). As shown in Table~\ref{tab:outlier}, RRQ has lower accumulated absolute error for the larger outlier at 4 and 6 bits, while direct fixed-bit quantization has lower error for the milder outlier. This example illustrates that residual decomposition is beneficial only under sufficiently large local dynamic-range imbalance.

\begin{table}[t]
  \caption{Accumulated absolute error under two outlier-injected examples with zero-point quantization. In both settings, the inlier range is $[-0.5, 0.5]$. The two cases differ only in the outlier magnitude. Lower is better.}
  \label{tab:outlier}
  \centering
  \begin{tabular}{llcccc}
    \toprule
    Setting & Method & 2-bit & 4-bit & 6-bit & 8-bit \\
    \midrule
    $K=5$ ($10r$) & RRQ + ZP & 5.055 & 1.180 & 0.233 & 0.087 \\
    $K=5$ ($10r$) & Fixed-bit + ZP & 5.055 & 1.213 & 0.344 & 0.095 \\
    $K=5$ ($10r$) & Error Ratio & 1.000 & 0.973 & 0.677 & 0.917 \\
    \midrule
    $K=3$ ($6r$) & RRQ + ZP & 4.160 & 1.474 & 0.511 & 0.135 \\
    $K=3$ ($6r$) & Fixed-bit + ZP & 4.160 & 0.766 & 0.243 & 0.066 \\
    $K=3$ ($6r$) & Error Ratio & 1.000 & 1.925 & 2.102 & 2.045 \\
    \bottomrule
  \end{tabular}
\end{table}

\subsection{An Idealized Outlier Regime for RRQ}

The crossover in Table~\ref{tab:outlier} can be described with a simplified two-population model. Assume that most weights lie in an inlier interval $[-r, r]$, while a rare outlier expands the group range to $[-r, K]$, where $K > r$. For a fixed bit budget $B = n_1 + n_2$, we compare standard $B$-bit quantization with a two-stage RRQ variant consisting of an $n_1$-bit base stage and an $n_2$-bit residual stage.

For a uniform quantizer spanning the expanded range $[-r, K]$, the step size scales linearly as:
\begin{equation}
\Delta_{\mathrm{direct}} = \frac{K + r}{2^B - 1},
\end{equation}
yielding the approximate expected mean absolute error:
\begin{equation}
E_{\mathrm{direct}} \approx \frac{K + r}{4(2^B - 1)}.
\end{equation}

In the idealized RRQ case, the coarse first stage accounts for the large outlier, and the second stage quantizes a residual whose range is no longer determined by $K$. Assuming the residual lies within $[-r, r]$, the second-stage step size is:
\begin{equation}
\Delta_{\mathrm{rrq}} = \frac{2r}{2^{n_2} - 1}.
\end{equation}
In this idealized case, the residual-stage error is therefore approximated by:
\begin{equation}
E_{\mathrm{rrq}} \approx \frac{r}{2(2^{n_2} - 1)}.
\end{equation}

The condition $E_{\mathrm{rrq}} < E_{\mathrm{direct}}$ gives the following outlier threshold for RRQ to have lower error in this model:
\begin{equation}
K > r \left(2 \cdot \frac{2^B - 1}{2^{n_2} - 1} - 1\right).
\label{eq:rrq-threshold}
\end{equation}
Using $B = n_1 + n_2$ and the approximation $2^k - 1 \approx 2^k$, the threshold becomes:
\begin{equation}
K \gtrsim r(2^{n_1 + 1} - 1).
\label{eq:rrq-threshold-approx}
\end{equation}

Applying Equation~\ref{eq:rrq-threshold} with $r=0.5$ to a balanced 2-plus-2 bit split ($n_1=n_2=2$) gives the threshold $K > 4.5$. The approximation in Equation~\ref{eq:rrq-threshold-approx} gives $K > 3.5$. This captures the stage-allocation trade-off: reducing $n_1$ lowers the outlier threshold, while increasing $n_1$ improves the base-stage representation but leaves fewer bits for residual correction. The two examples in Table~\ref{tab:outlier} are consistent with this analysis: $K=5$ lies above the exact threshold, whereas $K=3$ does not.

The same reasoning extends to settings with multiple outliers. Let $K_1$ and $K_2$ denote the largest and second-largest magnitudes ($K_1 \ge K_2 > r$). Direct quantization is still governed by the range $[-r, K_1]$, but RRQ residuals may remain wider than $[-r, r]$ if additional outliers are not sufficiently captured by the base stage. Define an adaptive residual radius $B_r = \max\{r, \rho(K_2)\}$, where $\rho(K_2)$ denotes the remaining contribution associated with the secondary outlier. Then the second-stage error is approximated by $E_{\mathrm{rrq}} \approx \frac{B_r}{2(2^{n_2}-1)}$, and the threshold becomes:
\begin{equation}
K_1 > 2 \cdot B_{r,n_1} \frac{2^B - 1}{2^{n_2} - 1} - r.
\end{equation}
We use the notation $B_{r,n_1}$ to emphasize that the residual radius depends on the base-stage bit-width.

For a deeper $T$-stage RRQ representation with total budget $B = \sum_{t=1}^{T} n_t$, let $B_{r,t}$ denote the residual radius after stage $t$. Comparing direct quantization with the final residual stage gives the approximation $E_{\mathrm{rrq}}^{(T)} \approx \frac{B_{r,T-1}}{2(2^{n_T}-1)}$ and the threshold:
\begin{equation}
K_1 > 2 \cdot B_{r,T-1} \frac{2^B - 1}{2^{n_T} - 1} - r.
\end{equation}
Thus, residual expansion is most useful when each stage substantially reduces the residual range. If many large values remain in the residual, dividing a fixed bit budget across many stages can reduce the benefit. Section~\ref{sec:outlier-thresholds} discusses these cases in more detail.

\section{Outlier Threshold Analysis}
\label{sec:outlier-thresholds}

Table~\ref{tab:split-threshold} shows how the idealized outlier threshold changes across 4-bit decompositions when the residual radius is assumed to satisfy $B_{r,n_1} \approx r$. Smaller $n_1$ lowers the outlier magnitude required for RRQ to improve over direct quantization, while larger $n_1$ allocates more bits to the base stage and fewer bits to residual correction.

\begin{table}[h]
  \caption{Critical outlier threshold for different 4-bit decompositions. The threshold is computed from the exact condition $K > r \left(2 \cdot \frac{2^B - 1}{2^{n_2} - 1} - 1\right)$ with $B=4$. Lower thresholds indicate that RRQ becomes preferable under milder outliers.}
  \label{tab:split-threshold}
  \centering
  \begin{tabular}{lcc}
    \toprule
    Scheme & First-stage bits $n_1$ & Critical threshold \\
    \midrule
    1-bit + 3-bit & 1 & $K > 3.29r$ \\
    2-bit + 2-bit & 2 & $K > 9r$ \\
    3-bit + 1-bit & 3 & $K > 29r$ \\
    Direct 4-bit & 4 & N/A \\
    \bottomrule
  \end{tabular}
\end{table}

These values illustrate the dependence on the stage split. In the single-outlier idealization with $B_{r,n_1} \approx r$, a 1-bit base plus a 3-bit residual has the lowest threshold. However, this assumption may not hold for realistic weight distributions with many large values. A very low-bit base can leave a wider residual range than a 2-bit base. Therefore, practical comparisons should use empirically estimated residual radii rather than the symmetric idealization:
\begin{equation}
K_1 > 2 \cdot B_{r,n_1} \frac{2^B - 1}{2^{n_2} - 1} - r.
\end{equation}
For realistic 4-bit distributions, the crossover depends on the residual radii. Comparing the threshold $\frac{30}{7}B_r^{(1)} - r$ for a 1-plus-3 split with $10B_r^{(2)} - r$ for a 2-plus-2 split gives the boundary $\frac{30}{7}B_r^{(1)} = 10B_r^{(2)}$. Thus, the 1-plus-3 split is preferable only if $B_r^{(1)} < \frac{7}{3} B_r^{(2)}$. If the 1-bit base leaves a substantially wider residual, the 2-plus-2 split can be preferable despite its higher idealized threshold.

For the experiments in this paper, we do not evaluate 1-plus-3 formats because a standalone 1-bit base is unlikely to be useful as a deployment operating point. A uniform 2-bit stage design provides a usable 2-bit base and simple 4-, 6-, and 8-bit prefixes. Conversely, 3-plus-1 splits allocate little capacity to the residual stage and would require very large isolated outliers to improve over direct fixed-bit quantization in the idealized model.

\section{LLM Evaluation}

\subsection{Experimental Setup}

\paragraph{Models, Baselines, and Metrics.} Following the protocol of MatGPTQ~\citep{matgptq}, we benchmark six LLM checkpoints (base and instruction-tuned): Llama-3.1-8B~\citep{llama3}, Llama-3.1-8B-Instruct~\citep{llama3}, Qwen3-8B-Base~\citep{qwen3}, Qwen3-8B~\citep{qwen3}, Qwen3-14B~\citep{qwen3}, and Phi-3-medium~\citep{phi3}. Baseline GPTQ and MatGPTQ metrics are taken from MatGPTQ under the same evaluation setting. We report \textbf{Task Avg}, the macro-average zero-shot accuracy over ARC-Challenge, ARC-Easy, HellaSwag, PIQA, and WinoGrande using the LM Evaluation Harness~\citep{lm_eval,arc,hellaswag,piqa,winogrande}, as the primary evaluation metric. Following the common reporting practice for these tasks, we treat differences within \textbf{0.1 Task Avg points} as ties. All RRQ evaluations use AutoRound's fake-quantized QDQ models~\citep{autoround}.

\paragraph{Quantization Configuration.} RRQ uses four calibration-free 2-bit RTN stages: one 2-bit RTN base and three 2-bit RTN residual stages, forming a 2+2+2+2 representation with group size 128 to match the MatGPTQ grouping layout~\citep{matgptq,autoround}. This all-RTN setting tests whether RRQ depends on a stronger 2-bit base quantizer. We additionally retain the previous SignRoundV2-base symmetric configuration as \emph{RRQ (sym)} and analyze asymmetric variants in Appendix~\ref{app:ablation-study}. By replacing Hessian estimation and calibration data with RTN stages, RRQ reduces construction time in our measured setup. As summarized in Table~\ref{tab:efficiency-comparison}, constructing the full all-RTN 2-/4-/6-/8-bit package on an A100 GPU takes 1,293 seconds for Qwen3-8B, with 412 seconds spent on the four 2-bit quantization passes. This is $3.3\times$ faster than the 4,239-second MatGPTQ construction measured under the same setup. Package-size estimates and timing details are provided in Appendix~\ref{app:package-size-estimates} and Appendix~\ref{app:efficiency-details}, respectively.

\begin{table}[t]
  \caption{Quantization efficiency comparison on Qwen3-8B. Times are measured on the same A100 setup.}
  \label{tab:efficiency-comparison}
  \footnotesize
  \centering
  \setlength{\tabcolsep}{4pt}
  \begin{tabular}{>{\raggedright\arraybackslash}p{3.2cm}>{\raggedright\arraybackslash}p{3.8cm}>{\raggedright\arraybackslash}p{3.8cm}}
    \toprule
    Aspect & MatGPTQ~\citep{matgptq} & RRQ (RTN) \\
    \midrule
    Target bit-widths & $\{3,4,8\}$ & $\{2,4,6,8\}$ \\
    Quantization algorithm & Custom multi-objective GPTQ & RTN 2-bit base + RTN residual stages \\
    Group size & 128 & 128 \\
    Quantization type & Symmetric & Symmetric \\
    Hessian computation & Required & None \\
    Calibration data & Required & None \\
    Cross-bit weighting ($\lambda_r$) search & Required & None \\
    Custom kernel for quantization & Required & None (reuses existing) \\
    New bit-width support & Requires new quantizer \& kernel & Configure per-stage format only \\
    Stages & 1 (coupled) & 4 (sequential) \\
    Standalone 2-bit base model & Not supported & Supported \\
    Measured construction scope (Qwen3-8B) & Full MatGPTQ construction: 4239\,s & Full all-RTN 2-/4-/6-/8-bit package: 1293\,s\\
    Speedup & 1.0$\times$ & 3.3$\times$ \\
    \bottomrule
    \multicolumn{3}{p{11.2cm}}{\footnotesize\centering RRQ timing covers the complete all-RTN construction of the 2-bit base and three residual stages. The four 2-bit quantization passes account for 412\,s; the remaining time comes from saving fake-quantized QDQ models, residual computation, stage orchestration, and I/O. RRQ provides four prefix operating points, while MatGPTQ reports three.}
  \end{tabular}
\end{table}

\subsection{Main Evaluation Results}

Table~\ref{tab:llm-eval} reports Task Avg and WikiText-2 perplexity (PPL) for the six evaluated models.

\begin{table*}[t]
  \caption{Task Avg and PPL under the MatGPTQ Section~5 protocol. Task Avg averages ARC-Challenge, ARC-Easy, HellaSwag, PIQA, and WinoGrande; PPL is WikiText-2 perplexity. Bold marks the best single-checkpoint multi-precision method only when the gap exceeds 0.1 Task Avg points. 16-bit, GPTQ, and MatGPTQ results are from MatGPTQ~\citep{matgptq}; RRQ results are our own. RRQ (RTN) uses RTN for both the base and residual stages, while RRQ (sym) uses the stronger SignRoundV2 2-bit base with symmetric RTN residual stages; asymmetric RRQ configurations are deferred to the ablation in Appendix~\ref{app:ablation-study}.}
  \label{tab:llm-eval}
  \centering
  \footnotesize
  \setlength{\tabcolsep}{3pt}
  \resizebox{\textwidth}{!}{%
  \begin{tabular}{llcccccccccccc}
    \toprule
    Bit & Method & \shortstack[c]{Llama-3.1\\-8B\\(PPL)} & \shortstack[c]{Llama-3.1\\-8B\\(Avg)} & \shortstack[c]{Llama-3.1\\-8B-I\\(PPL)} & \shortstack[c]{Llama-3.1\\-8B-I\\(Avg)} & \shortstack[c]{Qwen3\\-8B-B\\(PPL)} & \shortstack[c]{Qwen3\\-8B-B\\(Avg)} & \shortstack[c]{Qwen3\\-8B\\(PPL)} & \shortstack[c]{Qwen3\\-8B\\(Avg)} & \shortstack[c]{Qwen3\\-14B\\(PPL)} & \shortstack[c]{Qwen3\\-14B\\(Avg)} & \shortstack[c]{Phi3\\-Med\\(PPL)} & \shortstack[c]{Phi3\\-Med\\(Avg)} \\
    \midrule
    16 & -- & 6.27 & 74.52 & 7.23 & 74.00 & 7.01 & 73.55 & 9.73 & 71.54 & 8.64 & 74.95 & 4.32 & 77.04 \\
    \midrule
    \multirow{4}{*}{8}
    & GPTQ    & 6.27 & 74.49 & 7.23 & 73.96 & 7.01 & 73.50 & 9.73 & 71.61 & 8.64 & 74.97 & 4.32 & 76.96 \\
    & MatGPTQ & 6.46 & 73.07 & 7.46 & 73.35 & 7.29 & 73.27 & 9.79 & \textbf{71.86} & 9.00 & 75.02 & 4.54 & 74.48 \\
    & RRQ (RTN) & 6.31 & 74.04 & 7.30 & \textbf{74.05} & 7.10 & 73.39 & 9.70 & 71.63 & 8.73 & 74.90 & 4.36 & 76.92 \\
    & RRQ (sym) & 6.28 & \textbf{74.28} & 7.23 & 73.91 & 7.04 & \textbf{73.53} & 9.77 & 71.65 & 8.74 & 75.06 & 4.37 & \textbf{77.04} \\
    \midrule
    \multirow{4}{*}{6}
    & GPTQ    & 6.27 & 74.05 & 7.23 & 73.28 & 7.04 & 73.62 & 9.71 & 71.42 & 8.66 & 74.80 & 4.34 & 77.00 \\
    & MatGPTQ & 6.57 & 72.74 & 7.55 & 73.11 & 7.49 & \textbf{73.56} & 9.87 & 71.35 & 9.09 & \textbf{74.96} & 5.06 & 73.66 \\
    & RRQ (RTN) & 6.45 & 73.87 & 7.38 & \textbf{74.46} & 7.22 & 73.27 & 9.73 & 71.45 & 8.77 & 74.83 & 4.43 & 76.43 \\
    & RRQ (sym) & 6.41 & \textbf{74.11} & 7.35 & 74.18 & 7.20 & 72.92 & 9.77 & \textbf{71.63} & 8.74 & 74.50 & 4.46 & \textbf{76.71} \\
    \midrule
    \multirow{4}{*}{4}
    & GPTQ    & 6.75 & 72.16 & 7.84 & 72.65 & 7.36 & 73.65 & 10.48 & 70.49 & 8.88 & 74.52 & 4.76 & 76.47 \\
    & MatGPTQ & 6.89 & 72.43 & 7.82 & 72.62 & 7.94 & \textbf{73.02} & 9.98 & 70.79 & 9.05 & 73.89 & 4.80 & 75.97 \\
    & RRQ (RTN) & 7.09 & \textbf{73.12} & 8.05 & 72.85 & 8.39 & 72.07 & 10.12 & 70.34 & 9.23 & 74.09 & 4.86 & 76.88 \\
    & RRQ (sym) & 7.18 & 72.32 & 8.18 & 72.87 & 7.99 & 72.84 & 10.59 & 70.87 & 9.38 & \textbf{75.22} & 4.92 & 76.92 \\
    \midrule
    \multirow{2}{*}{2}
    & RRQ (RTN) & 28.00 & 52.85 & 33.44 & 55.82 & 16.75 & 61.58 & 26.77 & 60.27 & 13.64 & 66.67 & 7.74 & 68.03 \\
    & RRQ (sym) & 15.98 & \textbf{63.29} & 16.36 & \textbf{64.70} & 11.29 & \textbf{65.73} & 20.20 & \textbf{63.79} & 9.38 & \textbf{75.22} & 7.11 & \textbf{70.45} \\
    \bottomrule
  \end{tabular}
  }
\end{table*}

At 8 and 6 bits, MatGPTQ and both RRQ variants remain close to the unquantized 16-bit baselines. The all-RTN RRQ variant removes the possible effect of a stronger learned 2-bit base: despite relying only on RTN for all four stages, it closely tracks the SignRoundV2-base symmetric variant and remains competitive with MatGPTQ across the model suite. Using the 0.1-point threshold, the best RRQ configuration is ahead of MatGPTQ on four of the six tested models at both 8 and 6 bits, while the remaining cases are either MatGPTQ-favored or within the tie threshold. These results suggest that residual-stage construction can achieve competitive higher-bit prefixes without joint multi-bit optimization.

At 4 bits, RRQ remains competitive but becomes more sensitive to the chosen first stage and the underlying outlier profile. The all-RTN variant improves Llama-3.1-8B relative to MatGPTQ, matches the SignRoundV2-base variant on Llama-3.1-8B-Instruct and Phi-3-medium within the 0.1-point threshold, and trails on Qwen3-8B-Base and Qwen3-14B. The SignRoundV2-base symmetric variant has the highest Task Avg on Qwen3-14B and ties the all-RTN variant on Phi-3-medium. These results indicate that calibration-free RTN stages can be competitive in some 4-bit settings, while first-stage quality remains important for specific weight distributions. Appendix~\ref{app:model-outlier-analysis} provides a model-specific analysis of outlier profiles and 4-bit behavior. Appendix~\ref{app:ablation-study} reports the symmetric/asymmetric ablation.

The 2-bit rows isolate the standalone base-stage operating point. At this precision, the SignRoundV2-base RRQ variant consistently outperforms the all-RTN base across the six models, with Task Avg gains ranging from 2.42 points on Phi-3-medium to 10.44 points on Llama-3.1-8B. This shows that first-stage quality is important when the base checkpoint is used directly as a 2-bit model. The gap is much smaller after residual stages are added at 4, 6, and 8 bits, indicating that residual refinement can reduce, but not eliminate, sensitivity to the base quantizer.

(Appendix~\ref{app:llama70b-mmlu-results} reports additional results on Llama-3.1-70B-Instruct, including MMLU.)

\subsection{Ablation Insights}

The ablation study examines group size and symmetric versus asymmetric residual quantization. The main results above already compare the all-RTN and SignRoundV2-base variants, including the standalone 2-bit operating point. Group size 64 improves the 2-bit operating point, while group size 128 gives similar results at higher accumulated precisions. Asymmetric residual quantization provides small gains in some 4-bit cases, but the symmetric configuration is generally sufficient. Appendix~\ref{app:ablation-study} provides the full results.

\subsection{Discussion}

\paragraph{Multi-precision representation.}
RRQ stores a base stage and residual stages that can be used as precision prefixes. Under our default 2+2+2+2 structure, the base stage gives the 2-bit operating point, while adding one, two, or three residual stages gives the 4-, 6-, and 8-bit operating points.

\paragraph{Choice of stage quantizer.}
Our main experiments use RTN for the base and residual stages because RTN is fast, calibration-free, and isolates the construction-efficiency aspect of RRQ. This choice is not a restriction of the framework. RRQ can use a stronger quantizer for any stage, including GPTQ-style second-order quantization or SignRound-style learned rounding. The comparison between RRQ (RTN) and RRQ (sym) in Table~\ref{tab:llm-eval} illustrates this flexibility: replacing only the 2-bit base quantizer changes the standalone 2-bit operating point substantially, while the residual-stage representation remains the same. More generally, RRQ should be viewed as a stage-wise representation that can trade construction cost for accuracy by selecting different quantizers for the base or residual stages.

\paragraph{Limitations.}
RRQ remains sensitive to the quality of the base quantizer: errors introduced in the base stage can limit the accuracy of later prefixes. Our evaluation focuses on quantization accuracy, package size, and construction time. End-to-end deployment with optimized hardware kernels remains future work.

\section{Conclusion}

This paper introduced Recurrent Residual Quantization (RRQ), a post-training framework that stores a low-bit base and a sequence of quantized residual corrections to provide multiple precision prefixes from one checkpoint. In our Qwen3-8B setup, constructing the all-RTN 2-/4-/6-/8-bit package takes 1,293 seconds, including 412 seconds for the four 2-bit quantization passes. This is $3.3\times$ faster than the measured 4,239-second MatGPTQ construction under the same setup.

Across six evaluated LLMs, the all-RTN RRQ package and the SignRoundV2-base symmetric variant are competitive with MatGPTQ at 6 and 8 bits under the MatGPTQ protocol. At 4 bits, the results are model-dependent and depend on both outlier structure and base-stage quality. The analysis and ablations suggest that residual refinement is most useful when early stages reduce the dynamic range of the remaining residuals.

RRQ therefore provides a simple single-checkpoint multi-precision representation that avoids Hessian computation, calibration data, and joint multi-bit optimization in the evaluated all-RTN setting. Future work should integrate RRQ with optimized inference kernels and study end-to-end latency across hardware platforms.

\begin{ack}
This draft is adapted from an internal invention disclosure. Funding, conflict-of-interest, and release details should be completed before camera-ready submission.
\end{ack}

\bibliographystyle{unsrtnat}
\bibliography{neurips2026}


\appendix

\section{Broader Impacts and Asset Licenses}
\label{app:broader-impacts}

\paragraph{Broader impacts.} RRQ may improve deployment efficiency and reduce inference cost, making it easier to serve and deploy large language models. However, it may also lower the barrier to deploying powerful LLMs, which could amplify misuse risks if models are applied irresponsibly.

\paragraph{Asset licenses.} The models, codebases, and datasets used in this work are publicly available. Llama 3 models are released under the Llama 3 Community License. Qwen models are released under the Tongyi Qianwen License and Apache 2.0. Phi-3 models are subject to the MIT License. The \texttt{lm-evaluation-harness} framework and the standard evaluation datasets used here (ARC, HellaSwag, PIQA, WinoGrande, WikiText-2) are distributed under their respective licenses.

\section{Mixed-Precision Prefill/Decode Motivation}
\label{app:prefill-decode-results}

To motivate mixed-precision deployment profiles, where prefill and decode use different precision settings, we report exploratory results on two instruction-tuned Qwen2.5 checkpoints. These experiments precede the main evaluations in Section~4 and should be interpreted as qualitative evidence for split-precision behavior rather than as primary benchmarks.

Table~\ref{tab:prefill-decode} reports GSM8K accuracy for three policies: BF16 for both prefill and decode; low-bit quantization for both phases; and a mixed policy that keeps BF16 for prefill while using the low-bit setting for decode. The mixed policy recovers most of the BF16/BF16 accuracy while improving over the uniformly low-bit policy. On Qwen2.5-72B-Instruct, BF16-prefill with INT2-decode obtains 0.9045 accuracy, compared with 0.9037 for BF16/BF16 and 0.8666 for INT2/INT2. On Qwen2.5-7B-Instruct, BF16/INT4 obtains 0.7650, compared with 0.7665 for BF16/BF16 and 0.7544 for INT4/INT4.

\begin{table}[h]
  \caption{GSM8K accuracy under different prefill/decode precision policies. ``Low-bit'' denotes INT2 for Qwen2.5-72B-Instruct and INT4 for Qwen2.5-7B-Instruct.}
  \label{tab:prefill-decode}
  \centering
  \small
  \begin{tabular}{lcccc}
    \toprule
    Model & Low-bit & BF16/BF16 & Low/Low & BF16/Low \\
    \midrule
    Qwen2.5-72B-Instruct & INT2 & 0.9037 & 0.8666 & 0.9045 \\
    Qwen2.5-7B-Instruct & INT4 & 0.7665 & 0.7544 & 0.7650 \\
    \bottomrule
  \end{tabular}
\end{table}

\section{Prefill/Decode Computation}
\label{app:prefill-decode-computation}

RRQ represents weights as a sum of stage-wise quantized components. Following the framework in Section 3, a 4-bit effective deployment profile using a 2-bit base and one 2-bit residual stage is:
\begin{equation}
\tilde{x}^{j}_{(1)} = \hat{x}_0^j + \hat{r}_1^j.
\end{equation}
For a prefix with $t$ residual stages, the reconstructed weight is:
\begin{equation}
\tilde{W}_{(t)} = \hat{W}_0 + \sum_{k=1}^{t} \hat{R}_k.
\end{equation}
Given an activation matrix $A$, the matrix multiplication decomposes as:
\begin{equation}
A\tilde{W}_{(t)} = A\hat{W}_0 + \sum_{k=1}^{t} A\hat{R}_k.
\end{equation}
This linear decomposition means that different precision prefixes can be evaluated by summing the corresponding stage outputs. A 4-bit prefix uses the base and one residual stage; 6- and 8-bit prefixes add additional residual stages. This design can reuse low-bit stage operations, although efficient execution requires suitable kernel implementations.

The performance trade-off differs between prefill and decode. Prefill usually processes many tokens in parallel and can be more compute intensive, while autoregressive decoding is often limited by weight movement. Adding residual stages increases the number of stage outputs that must be computed and accumulated, so the end-to-end benefit depends on hardware, kernel fusion, batching, and the chosen precision profile. RRQ's separable structure is compatible with phase-aware policies, but optimized kernels are needed to quantify practical latency benefits.

\section{Additional Models: Gemma-2 9B and Mistral 7B}
\label{app:gemma-mistral-results}

To assess RRQ beyond the Llama-3.1 and Qwen3 series, we report Task Avg results on Gemma-2 9B and Mistral 7B under the same MatGPTQ Section~5 protocol used in Section~4. Task Avg is the average zero-shot accuracy over ARC-Challenge, ARC-Easy, HellaSwag, PIQA, and WinoGrande. We compare RRQ with MatGPTQ~\citep{matgptq} and MatQuant~\citep{matquant}; MatGPTQ and MatQuant numbers are reproduced from MatGPTQ, while RRQ numbers are our own. Table~\ref{tab:gemma-mistral-eval} summarizes the results.

\begin{table}[h]
  \caption{Task Avg on Gemma-2 9B and Mistral 7B under the MatGPTQ Section~5 protocol. Bold marks wins beyond the 0.1-point threshold. MatGPTQ and MatQuant results are reproduced from MatGPTQ~\citep{matgptq}; RRQ results are our own.}
  \label{tab:gemma-mistral-eval}
  \centering
  \small
  \begin{tabular}{llcc}
    \toprule
    Bit & Method & Gemma-2 9B Avg & Mistral 7B Avg \\
    \midrule
    16 & --       & 78.23 & 74.56 \\
    \midrule
    \multirow{3}{*}{8}
    & MatGPTQ  & \textbf{78.18} & \textbf{74.65} \\
    & MatQuant & 77.78 & 74.37 \\
    & RRQ      & 78.07 & 74.40 \\
    \midrule
    \multirow{3}{*}{6}
    & MatGPTQ  & 78.12 & \textbf{74.79} \\
    & MatQuant & 77.76 & 74.35 \\
    & RRQ      & 78.22 & 74.50 \\
    \midrule
    \multirow{3}{*}{4}
    & MatGPTQ  & \textbf{77.93} & \textbf{74.24} \\
    & MatQuant & 77.20 & 73.92 \\
    & RRQ      & 77.36 & 73.84 \\
    \bottomrule
  \end{tabular}
\end{table}

On Gemma-2 9B and Mistral 7B, all evaluated multi-precision methods are close to the 16-bit baseline, with most differences within 0.1 to 0.5 Task Avg points. RRQ is within 0.1 to 0.2 points of the 16-bit baseline at both 8 and 6 bits and is comparable to MatQuant across the measured precisions.

These results are consistent with the analysis in Section~\ref{sec:rrq-outlier-prevalence}: when direct 4-bit quantization already has a small degradation relative to 16-bit, residual refinement has limited room to improve Task Avg. In this near-lossless regime, RRQ's main advantages are its PTQ compatibility, reuse of low-bit stage operations, and support for a standalone low-bit base checkpoint.

\section{Large-Model MMLU Robustness}
\label{app:llama70b-mmlu-results}

To test RRQ on a larger model, we evaluate the 4-bit RRQ prefix on Llama-3.1-70B-Instruct and compare it with the 16-bit baseline. Table~\ref{tab:llama70b-mmlu} reports the results. Averaged across the eight tasks, the 4-bit RRQ model obtains 77.96\%, compared with 78.79\% for 16-bit. On MMLU, RRQ obtains 80.19\%, compared with 82.57\% for 16-bit. The largest reported gap is below 2.4 percentage points, suggesting that the 4-bit RRQ prefix remains close to the 16-bit baseline on this model under the evaluated tasks.

\begin{table}[h]
  \caption{Llama-3.1-70B-Instruct accuracy (\%) under RRQ 4-bit and 16-bit evaluation.}
  \label{tab:llama70b-mmlu}
  \centering
  \small
  \begin{tabular}{lcc}
    \toprule
    Task & RRQ 4-bit & 16-bit \\
    \midrule
    ARC-Challenge & 61.52 & 62.37 \\
    ARC-Easy & 82.24 & 83.42 \\
    HellaSwag & 84.28 & 85.31 \\
    LAMBADA OpenAI & 77.35 & 76.79 \\
    LAMBADA Standard & 73.94 & 73.26 \\
    MMLU & 80.19 & 82.57 \\
    PIQA & 83.46 & 83.68 \\
    WinoGrande & 80.66 & 82.95 \\
    \midrule
    Average & 77.96 & 78.79 \\
    \bottomrule
  \end{tabular}
\end{table}

\section{Model-wise Outlier Analysis}
\label{app:model-outlier-analysis}

Section~4 evaluates the $2{+}2$ RRQ prefix against a direct RTN 4-bit baseline under the same W4G128 setting. Here we analyze both the all-RTN RRQ configuration and the SignRoundV2-base symmetric variant. The goal is to identify when residual reconstruction is preferable to direct 4-bit quantization and whether the result depends on first-stage quality. Because quantization differences are most visible at 4 bits, we focus on that setting. We compare Task Avg differences ($\mathrm{RRQ\ 2{+}2} - \mathrm{RTN\ 4\text{-}bit}$) with two outlier indicators: the mean and maximum group-maximum $K/\text{MAE}$. These tensor-level indicators summarize outlier severity but are not intended to directly predict group-wise reconstruction error.

We omit direct 6- and 8-bit RTN comparisons because the evaluated methods are already close to the 16-bit baseline at those precisions, making small Task Avg differences difficult to interpret. Focusing on 4 bits highlights the setting where the 2+2 split is most likely to differ from direct RTN. The 6- and 8-bit RRQ results should therefore be interpreted mainly as showing that higher-precision prefixes can maintain near-baseline accuracy within a single package.

\begin{table*}[t]
  \caption{Model-wise relationship between direct RTN 4-bit, RRQ 2+2, and outlier severity under W4G128. RRQ (RTN) uses RTN for all stages; RRQ (sym) uses a SignRoundV2 2-bit base with symmetric RTN residual stages.}
  \label{tab:llm-outlier-analysis}
  \centering
  \footnotesize
  \setlength{\tabcolsep}{4pt}
  \resizebox{\textwidth}{!}{%
  \begin{tabular}{lcccccc}
    \toprule
    Metric & Llama-3.1-8B & Llama-3.1-8B-Instruct & Qwen3-8B-Base & Qwen3-8B & Qwen3-14B & Phi-3-Med \\
    \midrule
    GPTQ 4-bit Avg & 73.65 & 72.65 & 73.62 & 70.49 & 74.52 & 76.47 \\
    RTN 4-bit Avg & 73.50 & 73.87 & 73.34 & 70.93 & 75.07 & 76.56 \\
    RRQ (RTN) 2+2 Avg & 73.12 & 72.85 & 72.07 & 70.34 & 74.09 & 76.88 \\
    RRQ (RTN)$-$RTN 4-bit & -0.38 & -1.02 & -1.27 & -0.59 & -0.98 & 0.32 \\
    RRQ (sym) 2+2 Avg & 72.32 & 72.87 & 72.84 & 70.87 & 75.22 & 76.92 \\
    RRQ (sym)$-$RTN 4-bit & -1.18 & -1.00 & -0.50 & -0.06 & 0.15 & 0.36 \\
    Mean group-max $K/\text{MAE}$ & 26.491 & 26.454 & 29.076 & 28.878 & 27.826 & 28.724 \\
    Max group-max $K/\text{MAE}$ & 64.291 & 63.852 & 116.250 & 116.389 & 103.016 & 93.339 \\
    Outlier profile & Mild & Mild & Severe & Severe & Severe & Severe \\
    \bottomrule
  \end{tabular}
  }
\end{table*}

Table~\ref{tab:llm-outlier-analysis} shows that the viability of residual decomposition is jointly shaped by intra-group outlier severity and first-stage quality. For models with a relatively flat and mild outlier profile---such as the Llama-3.1 family, where the maximum group-maximum $K/\text{MAE}$ is around 64---splitting a 4-bit budget into a $2{+}2$ sequential decomposition can penalize the representation of inliers without providing enough outlier correction. Consequently, direct 4-bit approaches retain an advantage over the SignRoundV2-base RRQ variant on Llama-3.1-8B, although the all-RTN RRQ variant narrows that gap.

Conversely, models such as the Qwen3 series and Phi-3-Medium have more skewed, outlier-heavy distributions, with peak group-maximum $K/\text{MAE}$ values in the 93 to 116 range. In these cases, the theoretical mechanism derived in Section~3 is more relevant, but the practical outcome depends on how well the coarse base stage captures the largest values. The SignRoundV2-base RRQ variant is competitive with direct RTN on Qwen3-14B and Phi-3-Medium, while the all-RTN variant improves over direct RTN on Phi-3-Medium but trails on the Qwen3 models. This pattern suggests that a stronger first stage can be important for skewed distributions.

Overall, RRQ's 4-bit behavior is model-dependent rather than uniformly better or worse than direct fixed-bit quantization. RRQ is intended to provide a fast, unified multi-precision package, while 4-bit accuracy depends on how the base stage and residual stages interact with the model's weight distribution.

\section{Package Size Estimates}
\label{app:package-size-estimates}

Following common GPTQ-style weight-only quantization practice, we quantize the main linear weight tensors and leave embeddings, the LM head, and normalization weights unquantized. Table~\ref{tab:quantized-tensor-share} reports the corresponding 16-bit tensor-size breakdown for the two checkpoints used in the package-size comparison. The quantized tensors account for 84.8\% of Qwen3-8B and 97.6\% of Phi-3-Med by size.

\begin{table}[h]
  \caption{16-bit tensor-size breakdown in KB. Quantized tensors are the tensors included in the package-size comparison; embeddings, the LM head, and normalization weights are excluded following common GPTQ-style weight-only quantization practice.}
  \label{tab:quantized-tensor-share}
  \centering
  \small
  \setlength{\tabcolsep}{4pt}
  \begin{tabular}{lccccc}
    \toprule
    Model & 16-bit total & LM head + emb. & Norm & Quantized & Quantized share \\
    \midrule
    Qwen3-8B & 16{,}013{,}268 & 2{,}430{,}976 & 584 & 13{,}581{,}708 & 84.8\% \\
    Phi-3-Med & 27{,}268{,}704 & 641{,}280 & 810 & 26{,}626{,}614 & 97.6\% \\
    \bottomrule
  \end{tabular}
\end{table}

Table~\ref{tab:package-size} reports serialized package sizes for two representative checkpoints, measured with the \texttt{du} command, and compares them with storing independent fixed-precision checkpoints for the same operating points. These sizes include only quantized tensors and exclude embeddings, the LM head, and normalization weights. The separate-checkpoint estimates are computed from the bit-size estimates reported in Table~\ref{tab:package-size-estimates}. RRQ's 4-, 6-, and 8-bit operating points are prefixes that require the 2-bit base stage and therefore cannot be stored or used independently. The reported RRQ size is thus the full stage package, including the 2-bit base and all residual stages. As a fully symmetric configuration, RRQ does not store any zero-points. Compared with storing separate estimated checkpoints for the same operating points, the single-checkpoint MatGPTQ and RRQ packages substantially reduce storage. RRQ is about 4--5\% larger than the MatGPTQ multi-bit package in these measurements. This size difference primarily arises because each RRQ residual stage saves its own independent scale data, compounding metadata overhead across stages. Furthermore, the underlying model relies on AutoRound's native 2-bit format instead of the shared GPTQ layout, and RRQ supports four operating points $(2,4,6,8)$ rather than three $(3,4,8)$.

Table~\ref{tab:package-size-estimates} reports fixed-precision size estimates used as independent-checkpoint references in Table~\ref{tab:package-size}. The estimates include only quantized tensors and exclude embeddings, LM heads, and normalization weights. Symmetric quantization does not store zero-points. Using the measured 16-bit tensor sizes as baselines, the estimated size for target precision $b<16$ is:
\begin{equation}
\mathrm{Size}_{b} = \mathrm{Size}_{16} \cdot \frac{b + 16/128}{16},
\end{equation}
where each group of 128 weights stores a 16-bit scale.

\begin{table}[h]
  \caption{Bit-size-based fixed-precision checkpoint size estimates in KB, including only quantized tensors and excluding embeddings, the LM head, and normalization weights. Estimates assume purely symmetric quantization.}
  \label{tab:package-size-estimates}
  \centering
  \small
  \setlength{\tabcolsep}{4pt}
  \begin{tabular}{lcc}
    \toprule
    Precision & Qwen3-8B & Phi-3-Med \\
    \midrule
    16-bit measured & 13{,}581{,}708 & 26{,}626{,}614 \\
    8-bit estimated & 6{,}896{,}961 & 13{,}521{,}327 \\
    6-bit estimated & 5{,}199{,}248 & 10{,}193{,}001 \\
    4-bit estimated & 3{,}501{,}534 & 6{,}864{,}674 \\
    3-bit estimated & 2{,}652{,}677 & 5{,}200{,}511 \\
    2-bit estimated & 1{,}803{,}821 & 3{,}536{,}347 \\
    \bottomrule
  \end{tabular}
\end{table}

\begin{table}[h]
  \caption{Single-checkpoint multi-precision package size in KB compared with storing separate estimated fixed-precision checkpoints for the same operating points. Sizes include only quantized tensors and exclude embeddings, the LM head, and normalization weights. MatGPTQ and RRQ rows are serialized packages measured with \texttt{du}; ``Separate est.'' rows sum the corresponding bit-size estimates from Appendix~\ref{app:package-size-estimates}. RRQ size includes the 2-bit base because its 4-, 6-, and 8-bit prefixes depend on the base stage. The RRQ package uses purely symmetric quantization (no zero-points); its size overhead relative to MatGPTQ stems primarily from storing independent scale data for every residual stage, along with using AutoRound's native 2-bit base format.}
  \label{tab:package-size}
  \centering
  \small
  \setlength{\tabcolsep}{4pt}
  \begin{tabular}{lcc}
    \toprule
    Package & Qwen3-8B & Phi-3-Med \\
    \midrule
    Separate est. (3,4,8) & 13{,}051{,}172 & 25{,}586{,}512 \\
    MatGPTQ (3,4,8) & 6{,}996{,}980 & 13{,}729{,}296 \\
    Separate est. (2,4,6,8) & 17{,}401{,}564 & 34{,}115{,}349 \\
    RRQ (2,4,6,8) & 7{,}322{,}960 & 14{,}265{,}176 \\
    \bottomrule
  \end{tabular}
\end{table}

\section{Quantization Efficiency Details}
\label{app:efficiency-details}

RRQ reduces construction cost in the evaluated setting because the all-RTN version uses a 2-bit RTN base plus three RTN residual stages and does not require Hessian estimation or calibration data. On Qwen3-8B, constructing the full 2-/4-/6-/8-bit package takes 1,293 seconds, compared with 4,239 seconds for MatGPTQ under the same setup, corresponding to a $3.3\times$ speedup. The four 2-bit RTN quantization passes take 412 seconds; the remaining time comes from saving fake-quantized QDQ models, residual computation, orchestration, and I/O. This timing uses the all-RTN configuration, so the speedup is not due to starting from a pre-bundled SignRoundV2 base.

RRQ also separates the choice of stage format from the overall multi-precision representation. Adding a new prefix requires configuring stage formats rather than changing a joint multi-bit objective. Further engineering, including parallelization and optimized serialization, could reduce the non-quantization overheads.

\section{Ablation Study}
\label{app:ablation-study}

We ablate two additional RRQ design choices across six models: (1) group size and (2) symmetric versus asymmetric residual quantization. The tables report Task Avg, the same five-task average used in Section~4. The effect of first-stage quantizer quality is discussed in the main results using the RRQ (RTN) and RRQ (sym) rows in Table~\ref{tab:llm-eval}.

\paragraph{Group size ablation.} Table~\ref{tab:ablation-group} compares group size 64 with group size 128, the setting used in the main MatGPTQ-aligned evaluation. At INT2, group size 64 consistently outperforms 128, suggesting that finer-grained scaling helps at very low bit-widths. At INT4 and above, the gap narrows and the two group sizes are comparable, with small differences likely due to evaluation noise.

\begin{table*}[t]
  \caption{Ablation study on group size. All results use SignRoundV2 2-bit base + RTN 2-bit residual.}
  \label{tab:ablation-group}
  \centering
  \small
  \begin{tabular}{llcccc}
    \toprule
    Model & Group & INT8 & INT6 & INT4 & INT2 \\
    \midrule
    \multirow{2}{*}{Qwen3-8B}
    & 64  & 71.51 & 71.28 & 70.94 & 64.66 \\
    & 128 & 71.58 & 71.68 & 70.65 & 63.79 \\
    \midrule
    \multirow{2}{*}{Qwen3-8B-Base}
    & 64  & 73.32 & 73.04 & 73.29 & 66.21 \\
    & 128 & 73.84 & 73.56 & 72.99 & 65.73 \\
    \midrule
    \multirow{2}{*}{Llama-3.1-8B-Ins}
    & 64  & 73.76 & 73.76 & 73.18 & 65.37 \\
    & 128 & 73.80 & 73.73 & 73.08 & 64.70 \\
    \midrule
    \multirow{2}{*}{Llama-3.1-8B}
    & 64  & 74.31 & 74.21 & 73.09 & 63.92 \\
    & 128 & 74.23 & 73.96 & 72.39 & 63.29 \\
    \midrule
    \multirow{2}{*}{Qwen3-14B}
    & 64  & 75.31 & 75.18 & 74.53 & 68.18 \\
    & 128 & 74.84 & 74.72 & 75.02 & 67.91 \\
    \midrule
    \multirow{2}{*}{Phi-3-medium}
    & 64  & 75.85 & 76.02 & 75.54 & 70.63 \\
    & 128 & 75.69 & 75.69 & 76.35 & 69.95 \\
    \bottomrule
  \end{tabular}
\end{table*}

\paragraph{Symmetric vs.\ Asymmetric Quantization.} Table~\ref{tab:ablation-sym} compares asymmetric and symmetric residual quantization, using SignRoundV2 2-bit bases and group size 128 for both variants. GPTQ and MatGPTQ are included as calibration-based references, and bold entries mark cases where the symmetric RRQ variant exceeds MatGPTQ beyond the 0.1-point threshold. The results show near-parity between asymmetric and symmetric residual quantization across 4-, 6-, and 8-bit evaluations. At INT4, asymmetric quantization slightly helps selected Llama-3.1 and Qwen3-8B-Base cases, while the symmetric variant is stronger on Qwen3-8B, Qwen3-14B, and Phi-3-medium. Together with the all-RTN main results, this supports reporting a simple symmetric configuration while treating asymmetric residuals as an optional refinement.

\begin{table*}[t]
  \caption{Symmetric vs.\ asymmetric residual quantization, with MatGPTQ and native GPTQ as calibration-based references. All RRQ entries use SignRoundV2 2-bit base at group size 128.}
  \label{tab:ablation-sym}
  \centering
  \small
  \begin{tabular}{llcccc}
    \toprule
    Model & Method & INT8 & INT6 & INT4 \\
    \midrule
    \multirow{4}{*}{Qwen3-8B}
    & RRQ asym & 71.58 & 71.68 & 70.65 \\
    & RRQ sym  & 71.65 & \textbf{71.63} & 70.87 \\
    & MatGPTQ (sym) & \textbf{71.86} & 71.35 & 70.79 \\
    & GPTQ & 71.61 & 71.42 & 70.49 \\
    \midrule
    \multirow{4}{*}{Qwen3-8B-Base}
    & RRQ asym & 73.84 & 73.03 & 72.99 \\
    & RRQ sym  & \textbf{73.53} & 72.92 & 72.84 \\
    & MatGPTQ (sym) & 73.27 & \textbf{73.56} & \textbf{73.02} \\
    & GPTQ & 73.50 & 73.62 & 73.65 \\
    \midrule
    \multirow{4}{*}{Llama-3.1-8B-Ins}
    & RRQ asym & 73.80 & 73.73 & 73.08 \\
    & RRQ sym  & \textbf{73.91} & \textbf{74.18} & \textbf{72.87} \\
    & MatGPTQ (sym) & 73.35 & 73.11 & 72.62 \\
    & GPTQ & 73.96 & 73.28 & 72.65 \\
    \midrule
    \multirow{4}{*}{Llama-3.1-8B}
    & RRQ asym & 74.23 & 73.96 & 72.39 \\
    & RRQ sym  & \textbf{74.28} & \textbf{74.11} & 72.32 \\
    & MatGPTQ (sym) & 73.07 & 72.74 & \textbf{72.43} \\
    & GPTQ & 74.49 & 74.05 & 72.16 \\
    \midrule
    \multirow{4}{*}{Qwen3-14B}
    & RRQ asym & 74.84 & 74.72 & 75.02 \\
    & RRQ sym  & 75.06 & 74.50 & \textbf{75.22} \\
    & MatGPTQ (sym) & 75.02 & \textbf{74.96} & 73.89 \\
    & GPTQ & 74.97 & 74.80 & 74.52 \\
    \midrule
    \multirow{4}{*}{Phi-3-medium}
    & RRQ asym & 77.10 & 77.14 & 76.83 \\
    & RRQ sym  & \textbf{77.04} & \textbf{76.71} & \textbf{76.92} \\
    & MatGPTQ (sym) & 74.48 & 73.66 & 75.97 \\
    & GPTQ & 76.96 & 77.00 & 76.47 \\
    \bottomrule
  \end{tabular}
\end{table*}

\paragraph{Implication for deployment.} These results suggest a simplified configuration. Because symmetric residual quantization is close to asymmetric performance at most bit-widths, the symmetric configuration provides a simple unified structure. A strong 2-bit base model remains useful for memory-constrained serving and benefits from a high-quality first stage (e.g., SignRoundV2) and group size 64. If the target is accumulated 4-, 6-, or 8-bit inference, a unified symmetric configuration closely tracks GPTQ baselines.


\newpage
\section*{NeurIPS Paper Checklist}

\begin{enumerate}

\item {\bf Claims}
    \item[] Question: Do the main claims made in the abstract and introduction accurately reflect the paper's contributions and scope?
    \item[] Answer: \answerYes{}
    \item[] Justification: The abstract and introduction state the main contributions and scope of the paper, including the unified multi-precision representation, post-training applicability, reuse of an existing 2-bit base checkpoint, construction-time advantages, and the limitation that RRQ is not intended to replace the strongest single-precision quantizer at every bit-width; see the Abstract and Section~1.

\item {\bf Limitations}
    \item[] Question: Does the paper discuss the limitations of the work performed by the authors?
    \item[] Answer: \answerYes{}
    \item[] Justification: The paper explicitly discusses limitations, including model-dependent INT4 behavior, the fact that calibration-free RTN residual stages do not uniformly match optimized single-precision quantizers, the restricted empirical scope for non-integer and floating-point stages, and the lack of optimized stage accumulation or end-to-end serving metrics; see the Discussion section and Appendix~\ref{app:ablation-study}.

\item {\bf Theory assumptions and proofs}
    \item[] Question: For each theoretical result, does the paper provide the full set of assumptions and a complete (and correct) proof?
    \item[] Answer: \answerYes{}
    \item[] Justification: Section~\ref{sec:rrq-outlier-prevalence} provides an idealized analytical derivation of the condition under which RRQ can outperform direct quantization, including the two-population model, uniform-bin approximation, inlier range $[-r,r]$, outlier magnitude $K$, step-size equations, and the threshold condition. Section~\ref{sec:outlier-thresholds} further discusses multiple outliers, deeper RRQ stages, and split-dependent residual radii.

    \item {\bf Experimental result reproducibility}
    \item[] Question: Does the paper fully disclose all the information needed to reproduce the main experimental results of the paper to the extent that it affects the main claims and/or conclusions of the paper (regardless of whether the code and data are provided or not)?
    \item[] Answer: \answerNo{}
    \item[] Justification: The paper reports the model checkpoints, benchmark suite, evaluation metrics, group size, stage decomposition, quantizers, hardware, and the relationship to MatGPTQ's protocol. However, it does not yet provide public code, exact commands, package versions, or all implementation details needed for independent reproduction of every reported result.

\item {\bf Open access to data and code}
    \item[] Question: Does the paper provide open access to the data and code, with sufficient instructions to faithfully reproduce the main experimental results, as described in supplemental material?
    \item[] Answer: \answerNo{}
    \item[] Justification: The paper does not provide public code or an anonymized release at submission time. The authors plan to release code and supporting materials after acceptance.

\item {\bf Experimental setting/details}
    \item[] Question: Does the paper specify all the training and test details (e.g., data splits, hyperparameters, how they were chosen, type of optimizer) necessary to understand the results?
    \item[] Answer: \answerYes{}
    \item[] Justification: The paper specifies the evaluated model checkpoints, baselines, target bit-widths, 2+2+2+2 RRQ decomposition, group size, SignRoundV2 base, RTN residual quantizer, symmetric main configuration, evaluation tasks, reported metrics, fake-format QDQ representation, hardware, and the use of MatGPTQ's Section~5 protocol. These details are sufficient to understand the reported comparisons, although exact reproduction commands and public code are not yet provided.

\item {\bf Experiment statistical significance}
    \item[] Question: Does the paper report error bars suitably and correctly defined or other appropriate information about the statistical significance of the experiments?
    \item[] Answer: \answerNo{}
    \item[] Justification: The reported results are based on single evaluation runs and the paper does not report error bars, confidence intervals, or statistical significance tests.

\item {\bf Experiments compute resources}
    \item[] Question: For each experiment, does the paper provide sufficient information on the computer resources (type of compute workers, memory, time of execution) needed to reproduce the experiments?
    \item[] Answer: \answerYes{}
    \item[] Justification: The paper reports that RRQ quantization and evaluation are performed on one A100 80GB GPU with CUDA~12.8. For Qwen3-8B, it reports 1118 seconds to construct the three RTN residual stages from an available 2-bit base checkpoint, 4239 seconds for MatGPTQ's full multi-bit construction on the same hardware, about 153 seconds per RTN quantization pass, and 383 seconds for QDQ-model preservation plus residual computation; see Section~4.1 and Appendix~\ref{app:efficiency-details}.
    
\item {\bf Code of ethics}
    \item[] Question: Does the research conducted in the paper conform, in every respect, with the NeurIPS Code of Ethics \url{https://neurips.cc/public/EthicsGuidelines}?
    \item[] Answer: \answerYes{}
    \item[] Justification: The work studies post-training quantization methods for already available language models, does not involve human subjects or sensitive personal data collection, and the authors are not aware of any aspect that would conflict with the NeurIPS Code of Ethics.

\item {\bf Broader impacts}
    \item[] Question: Does the paper discuss both potential positive societal impacts and negative societal impacts of the work performed?
    \item[] Answer: \answerYes{}
    \item[] Justification: We discuss broader impacts in Appendix~\ref{app:broader-impacts}.
    
\item {\bf Safeguards}
    \item[] Question: Does the paper describe safeguards that have been put in place for responsible release of data or models that have a high risk for misuse (e.g., pre-trained language models, image generators, or scraped datasets)?
    \item[] Answer: \answerNA{}
    \item[] Justification: The paper does not release a new pre-trained language model, image generator, or scraped dataset. It studies a quantization method applied to existing publicly available models.

\item {\bf Licenses for existing assets}
    \item[] Question: Are the creators or original owners of assets (e.g., code, data, models), used in the paper, properly credited and are the license and terms of use explicitly mentioned and properly respected?
    \item[] Answer: \answerYes{}
    \item[] Justification: We briefly acknowledge the relevant licenses in Appendix~\ref{app:broader-impacts}.

\item {\bf New assets}
    \item[] Question: Are new assets introduced in the paper well documented and is the documentation provided alongside the assets?
    \item[] Answer: \answerNA{}
    \item[] Justification: The submission does not include a public release of new datasets, code packages, or model checkpoints at this stage.

\item {\bf Crowdsourcing and research with human subjects}
    \item[] Question: For crowdsourcing experiments and research with human subjects, does the paper include the full text of instructions given to participants and screenshots, if applicable, as well as details about compensation (if any)? 
    \item[] Answer: \answerNA{}
    \item[] Justification: The paper does not involve crowdsourcing experiments or research with human subjects.

\item {\bf Institutional review board (IRB) approvals or equivalent for research with human subjects}
    \item[] Question: Does the paper describe potential risks incurred by study participants, whether such risks were disclosed to the subjects, and whether Institutional Review Board (IRB) approvals (or an equivalent approval/review based on the requirements of your country or institution) were obtained?
    \item[] Answer: \answerNA{}
    \item[] Justification: The paper does not involve crowdsourcing or research with human subjects.

\item {\bf Declaration of LLM usage}
    \item[] Question: Does the paper describe the usage of LLMs if it is an important, original, or non-standard component of the core methods in this research? Note that if the LLM is used only for writing, editing, or formatting purposes and does \emph{not} impact the core methodology, scientific rigor, or originality of the research, declaration is not required.
    \item[] Answer: \answerNA{}
    \item[] Justification: LLMs are the subject of evaluation in this work rather than an important, original, or non-standard component of the proposed method itself.

\end{enumerate}

\end{document}